\pdfoutput=1
\documentclass[leqno,onefignum,onetabnum]{siamltex}
\usepackage[square,sort,comma,numbers]{natbib}

\usepackage[utf8]{inputenc} % allow utf-8 input
\usepackage[T1]{fontenc}    % use 8-bit T1 fonts
\usepackage{url}            % simple URL typesetting
\usepackage{booktabs}       % professional-quality tables
\usepackage{amsfonts}       % blackboard math symbols
\usepackage{nicefrac}       % compact symbols for 1/2, etc.
\usepackage{microtype}      % microtypography
\usepackage{soul}
\usepackage[utf8]{inputenc}
\usepackage{graphicx}
\usepackage{booktabs}

\usepackage[left=2.5cm,right=2.5cm, top=2.2cm, bottom=2.2cm]{geometry}

\usepackage[colorlinks=true,
citecolor=blue,
filecolor=black,
linkcolor=blue,
urlcolor=blue]{hyperref}

\usepackage{microtype}
\usepackage{graphicx}
\usepackage{subfigure}
\usepackage{booktabs} % for professional tables

\usepackage{multirow}
\usepackage{paralist}

\usepackage{booktabs} % for professional tables
\usepackage{multicol}
\usepackage{diagbox}

\usepackage[ruled,noend]{algorithm2e}

\SetCommentSty{mycommfont}

\usepackage{here}

\usepackage{amsmath,amssymb,amsfonts,amsbsy,amsfonts,latexsym}
\usepackage{multirow}
\usepackage{makecell}
\usepackage[labelfont=bf,textfont=it,belowskip=0pt,aboveskip=5pt,tableposition=top]{caption}
\usepackage{xcolor}
\usepackage{colortbl}

\definecolor{colorA}{RGB}{189,201,225}
\definecolor{colorB}{RGB}{103,169,207}
\definecolor{colorC}{RGB}{ 28,144,153}
\definecolor{colorD}{RGB}{  1,108, 89}

\newcolumntype{R}{>{\columncolor{gray!40}}r}
\newcolumntype{L}{>{\columncolor{gray!40}}l}
\newcolumntype{C}{>{\columncolor{gray!40}}c}

\usepackage{tabularx,colortbl,xcolor}
\usepackage{multirow}
\usepackage[normalem]{ulem}
\useunder{\uline}{\ul}{}

\usepackage{enumitem}

\usepackage{xparse}% http://ctan.org/pkg/xparse

\DeclareGraphicsExtensions{.pdf,.png}

\SetKwInput{KwInput}{Input}

\usepackage{longtable}
\usepackage{pgfplots}

\usepackage{amsmath}

\usepackage{xcolor}
\usepackage{colortbl}

\def\1{\bm{1}}

\usepackage[utf8]{inputenc}
\usepackage[english]{babel}

\newcommand\norm[1]{\left\lVert#1\right\rVert}

\newtheorem{theorem}{Theorem}
\newtheorem{assumption}{Assumption}

\newtheorem{lemma}[theorem]{Lemma}
\newenvironment{proof}{\paragraph{Proof}}{\hfill$\square$}

\DeclareMathOperator{\E}{\mathbb{E}}

\NewDocumentCommand{\var}{O{s} m O{}}{%
  \ensuremath{#1_{#2}^{#3}}% add \vphantom{<bizarre sup>}
}
\usepackage{siunitx}

\definecolor{light-gray}{gray}{0.80}

\newcommand\fref{Figure~\ref}
\newcommand\tref{Table~\ref}

\newcommand\ha{ \rowcolor{orange!0}}

\newcommand\hc{ \rowcolor{orange!40}}

\def\0{{\bf 0}}

\def\R{{\mathbb R}}

\newcommand{\MP}{\xspace\tiny{MP}}

\newcommand{\OURS}{\textsc{HAWQ-V2}\xspace}
\newcommand{\PREV}{\textsc{HAWQ}\xspace}

\usepackage{fancyhdr}
\title{\large{HAWQ-V2: Hessian Aware trace-Weighted Quantization of Neural Networks}}

\author{
  \textbf{\normalsize
Zhen Dong$^1$, Zhewei Yao$^1$,  Yaohui Cai\footnote{Equal Contribution}$^{\ ,2}$, Daiyaan Arfeen$^{*,1}$\\
Amir Gholami$^1$, Michael W. Mahoney$^1$, Kurt Keutzer$^1$}\\
\normalsize
$^1$University of California, Berkeley, \
$^2$Peking University\\
{\tt\small \{zhendong, zheweiy, daiyaanarfeen, amirgh, mahoneymw, and keutzer\}@berkeley.edu  caiyaohui@pku.edu.cn } 
}

\begin{document}

\maketitle
\thispagestyle{empty}

%%%%%%%% BODY TEXT

\begin{abstract}
\normalsize
Quantization is an effective method for reducing memory footprint and inference time of Neural Networks, e.g., for efficient inference in the cloud, especially at the edge.
However, ultra low precision quantization could lead to significant degradation in model generalization. 
A promising method to address this is to perform mixed-precision quantization, where more sensitive layers are kept at higher precision. 
However, the search space for a mixed-precision quantization is exponential in the number of layers.
Recent work has proposed a novel Hessian based framework~\cite{dong2019hawq}, with the aim of reducing this exponential search space by using second-order information.
While promising, this prior work has three major limitations: 
(i) they only use the top Hessian eigenvalue as a measure of sensitivity and do not consider the rest of the Hessian spectrum;
(ii) their approach only provides relative sensitivity of different layers and therefore requires a manual selection of the mixed-precision setting; and
(iii) they do not consider mixed-precision activation quantization.
Here, we present~\OURS which addresses these shortcomings.
For (i), we perform a theoretical analysis showing that a better sensitivity metric is to compute the average of all of the Hessian eigenvalues.
For (ii), we develop a Pareto frontier based method for selecting the exact bit precision of different layers without any manual selection.
For (iii), we extend the Hessian analysis to mixed-precision activation quantization. 
We have found this to be very beneficial for object detection.
We show that~\OURS achieves new state-of-the-art results for a wide range of tasks. In particular, 
we present quantization results for Inception-V3 (7.57MB with $75.68\%$ accuracy), ResNet50 (7.99MB with $75.76\%$ accuracy), and SqueezeNext (1MB with $68.38\%$ accuracy), all without any manual bit selection.
Furthermore, we present results for object detection on Microsoft COCO dataset, where we achieve 2.6 higher mAP than direct uniform quantization and 1.6 higher mAP than the recently proposed method of FQN, with an even smaller model size of 17.9MB.
\end{abstract}

\section{Introduction}
\label{sec:intro}

Deep convolutional Neural Networks (NNs) have achieved great success in recent years.
However, many of these models, particularly those with state-of-the-art performance, have a high computational cost and memory footprint. 
This slows inference and training in the cloud, and it prohibits their deployment on edge devices.

Quantization~\cite{asanovic1991experimental, courbariaux2015binaryconnect, rastegari2016xnor, zhou2017incremental, zhou2016dorefa, choi2018pact, zhang_2018_lqnets, dong2019hawq} is a very promising approach to address this problem by reducing the memory bottleneck, thus allowing the use of lower precision computational units in hardware.
By replacing floating point weights in the model with low precision fixed-point values, quantization can shrink the model size without changing the original network architecture. 
Moreover, in the case where both weights and activations are quantized to low precision, the expensive floating point matrix multiplication between weights and activations can be efficiently implemented using low-precision arithmetic with simpler operands and operators.
This can significantly reduce the inference latency on embedded~platforms.
The gains in speed and power consumption directly depend on how aggressively we can perform quantization without losing generalization/accuracy of the model.
Despite significant advances, performing ultra low-bit quantization results in significant degradation in accuracy.

\begin{figure}[t]
\centering
\includegraphics[width=.98\textwidth]{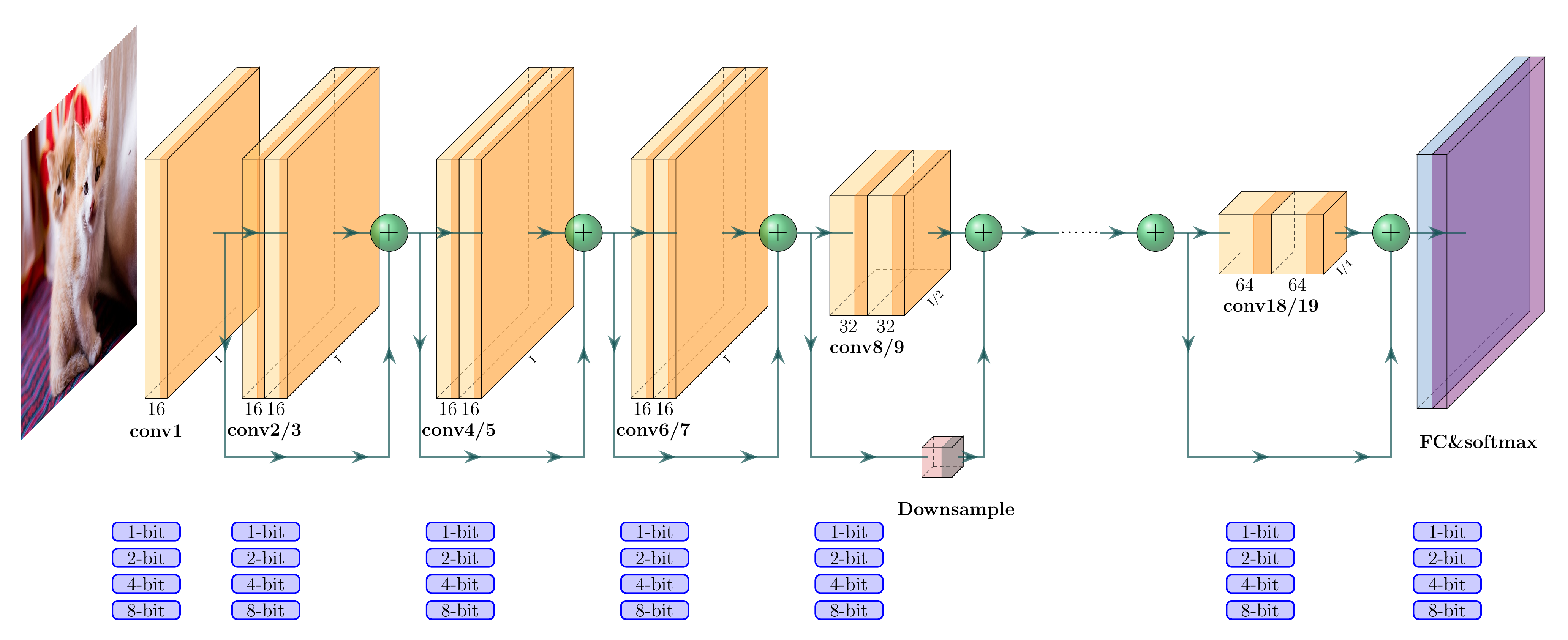}
\caption{
  Mixed Precision Illustration of ResNet20. Here we show the network architecture and list four possible bit precision setting for each layer. Since the number of possible bit settings is an exponential function of the number of blocks in a given network, we propose \OURS to generate precision settings automatically based on Hessian information instead of using simple search methods~\cite{wu2018mixed, wang2018haq}.
}
  \label{fig:resnet20_mixed_precision}
\end{figure}

Notable recent work on quantization includes 
using non-uniform quantizers~\cite{zhang_2018_lqnets},  
channel-wise~\cite{krishnamoorthi2018whitepaper} or group-wise quantization~\cite{shen2019Q-BERT} for weights, 
progressive quantization-aware fine-tuning~\cite{zhou2017incremental, dong2019hawq}, and 
mixed-precision quantization~\cite{wu2016quantized, wang2018haq, dong2019hawq}.
Despite the use of non-uniform quantization (which is generally difficult for efficient implementation in hardware), the accuracy degradation is still significant for ultra-low precision quantization.
A promising approach to address this is through mixed-precision quantization, where some layers are kept at higher precision, and other layers at lower precision.
However, a major problem with this approach is that the search space for determining a good mixed-precision quantization setting is exponentially large in the number of NN layers. 
This is schematically shown in~\fref{fig:resnet20_mixed_precision}, where we have assumed 4 precision options of 1/2/4/8 bits for each layer in a ResNet20 model.
Finding a mixed-precision setting using these bit precision, has a search space of size $4^{20} \approx 1\times 10^{12}$ (four times larger than the number of stars in the Milky Way). 
It is computationally impossible to test all of these mixed-precision settings and choose a particular setting with good generalization and hardware performance (in terms of latency and power).
Some recent work has proposed a reinforcement learning based method~\cite{wang2018haq} to address this. 
Another notable approach is differentiable neural architecture search (DNAS) based methods~\cite{wu2016quantized}.
However, these searching methods can require a large amount of computational resources, are time-consuming, and, worst of all, the quality of quantization is very sensitive to the initialization of their search parameters and therefore unpredictable. 
This makes deployment of these methods in online learning scenarios especially challenging, as in these applications a new model is trained every few hours and needs to be quantized for efficient inference.

To address these issues, recent work introduced~\PREV \cite{dong2019hawq}, a Hessian AWare Quantization framework. 
The main idea is to assign higher bit-precision to layers that are \emph{more sensitive}, and lower bit-precision to \emph{less sensitive} layers. 
This sensitivity is measured through second-order information, as computed via the Hessian operator. 
In particular, \PREV computes the top Hessian eigenvalue of each layer and uses this as a metric to sort the sensitivity of different layers. 
This can significantly reduce the exponential search space for mixed-precision quantization, since a layer with higher Hessian eigenvalues cannot be assigned lower bits, as compared to another layer with smaller Hessian eigenvalues.
However, there are several shortcomings of this approach: 
(i) \PREV only uses the top Hessian eigenvalue as a measure of sensitivity, and it ignores the rest of the Hessian spectrum;
(ii) \PREV only provides relative sensitivity of different layers, and it still requires a manual selection of the mixed-precision setting; and
(iii) \PREV does not consider mixed-precision activation quantization.

%%%%%%%%%%%%%%%%%%%%%%%%%%%%%%%%%%%%%%%%%%%%%%%%%%%%%%%%%%%%%%%%%%%%%%%%%%%%%%%%%%%
%%%%%%%%%%%%%%%%%%%%%%%%%%%%%%%%%%%%%%%%%%%%%%%%%%%%%%%%%%%%%%%%%%%%%%%%%%%%%%%%%%%

Here, we address these challenges and introduce the \OURS method.
Our contributions are as follows.

\begin{enumerate}
    \item 
    We perform a theoretical analysis (Lemma~\ref{lemma_1}) showing that a better sensitivity metric is to use the average Hessian trace, instead of just the top eigenvalue as used in \PREV~\cite{dong2019hawq}.
    \item 
    We implement a fast algorithm to compute Hessian trace information using Hutchinson's algorithm in PyTorch. 
    (Recall that the trace of a square matrix is the sum of the elements along the main diagonal.)
    For example, we can compute Hessian trace for all 54 layers in ResNet50 in 30 minutes with 4 GPUs (only 33s per block on average). 
    A common concern with the application of Hessian-based methods is the computational cost, but we demonstrate that (when implemented properly) this is not an issue.
    \item 
    The \PREV framework~\cite{dong2019hawq} only provides relative sensitivity, and thus it requires that the precise bit-precision setting is manually determined. 
    We address this by using a Pareto-frontier based method to determine automatically the bit precision of different layers without any manual selection (\fref{fig:omega_size}).
    \item 
    We extensively test \OURS for a wide range of problems, and we achieve new state-of-the-art results. 
    In particular, we present quantization results for Inception-V3 (\tref{tab:inception_compress_ratio}), ResNet50 (\tref{tab:resnet50_compress_ratio}), and SqueezeNext (\tref{tab:squeezenext_compress_ratio}).
    Furthermore, we present results for object detection on the Microsoft COCO dataset, where \OURS achieves 2.6 higher mAP than direct uniform quantization and 1.6 higher mAP than the recently proposed method of FQN~\cite{li2019fully}, with even smaller model size 17.9MB (\tref{tab:retinanet}).
    \item 
    We extend the \PREV work~\cite{dong2019hawq} to mixed-precision activation quantization, as described in \S\ref{activation_trace}. 
    We propose a fast method for computing Hessian information w.r.t. activations, and we show that mixed-precision activation can boost the performance of the object detection model mentioned before to 34.4 mAP (\tref{tab:retinanet}).
\end{enumerate}

\emph{Outline:} 
In \S~\ref{sec:methodology}, we discuss theoretical analysis and the relationship between the Hessian spectrum and quantization. 
We then discuss the Pareto frontier and our automatic precision selection method.
Then, in \S~\ref{sec:results}, we show the results of the trade-off between speed and convergence in the Hutchinson algorithm; and we test \OURS with various models on both image classification and object detection tasks. 
Finally, in \S~\ref{sec:conclusions}, we provide a brief conclusion and discussion of future work.

\section{Methodology}
\label{sec:methodology}

For a supervised learning framework, the goal is to minimize the empirical risk loss,
\begin{equation}
    \mathcal{L}(\theta) = \frac{1}{N}\sum_{i=1}^N f(x_i, y_i, \theta),
\end{equation}
where $\theta\in R^d$ is the learnable model parameters, and $f(x,y,\theta)$ is the loss for a datum $(x,y)\in(X,Y)$. 
Here, $X$ is the input, $Y$ is the corresponding label, and $N=|X|$ is the cardinality of the training set. 
Assume that the NN can be partitioned into $L$ layers as $\{B_1, B_2, \cdots, B_L\}$, with corresponding learnable parameters $\{W_1, W_2, \cdots, W_L\}$.
Furthermore, we denote mini-batch gradient of the loss w.r.t. model parameters as $g=\frac{1}{N_B}\sum_{i=1}^{N_B}\frac{\partial f}{\partial \theta}$, and sub-sampled
Hessian w.r.t. model parameters as $H=\frac{1}{N_B}\sum_{i=1}^{N_B}\frac{\partial^2 f}{\partial \theta^2}$, where $N_B$ is the mini-batch.

For quantization, we assume that the model is trained and all of its weights and activations are stored in single precision (32-bit).
To reduce the memory footprint and inference time, we quantize the weights and activations by restricting
their values
to a finite set of numbers, using the following quantization function:
\begin{equation}
    Q(z) = q_j,~~~~~~\text{for}~z\in(t_j, t_{j+1}],
\end{equation}
where $(t_j, t_{j+1}]$ denotes an interval in the real numbers ($j=0,~\ldots~,2^k-1$), $k$ is the quantization precision, and $z$ stands for either activations or weights.

This is a non-differentiable function and typically can be addressed by using the Straight Through Estimator (STE)~\cite{bengio2013estimating} to backpropagate the gradients. 
See Appendix~\ref{sec:quantization} for details.

As mentioned before, using ultra-low bit precision for $Q$ for all layers can lead to significant accuracy loss. 
A viable method to address this is to use mixed-precision quantization, where \textit{more sensitive} layers are kept at \emph{higher precision}. 
However, as mentioned before, the search space for mixed-precision quantization
is exponential in the number of layers. Below we perform a theoretical analysis to find a sensitivity
metric to eliminate searching through this exponentially large set.

\subsection{Trace Weighted Quantization}
In the previous \PREV work~\cite{dong2019hawq}, the top eigenvalue of the Hessian was used to determine the relative sensitivity order of different layers. 
However, a NN model contains millions of parameters, and thus millions of Hessian eigenvalues. 
Therefore, just measuring the top eigenvalue may be sub-optimal.
As a simple example, consider two functions $F_1(x,y) = 100x^2+y^2$ and $F_2(x,y)= 100x^2+99y^2$. 
The top Hessian eigenvalues of $F_1$ and $F_2$ are the same (i.e., 200). 
However, it is clear that $F_2$ is more sensitive than $F_1$ since $F_2$ has much larger function value change along y-axis. 
Below, we perform a theoretical analysis and show that a better metric is to compute the average Hessian trace (i.e., average of all Hessian eigenvalues) instead of just the top eigenvalue.

%%%%%%%%%%%%%%%%%%%%%%%%%%%%%%%%%%%%%%%%%%%%%%%%%%%%%%%%%%%%%%%%%%%%%%%%%%%%%%%%%
\begin{figure}[t]
\centering
\includegraphics[width=0.45\textwidth,trim=0.2in 0in 0.in 0.in, clip]{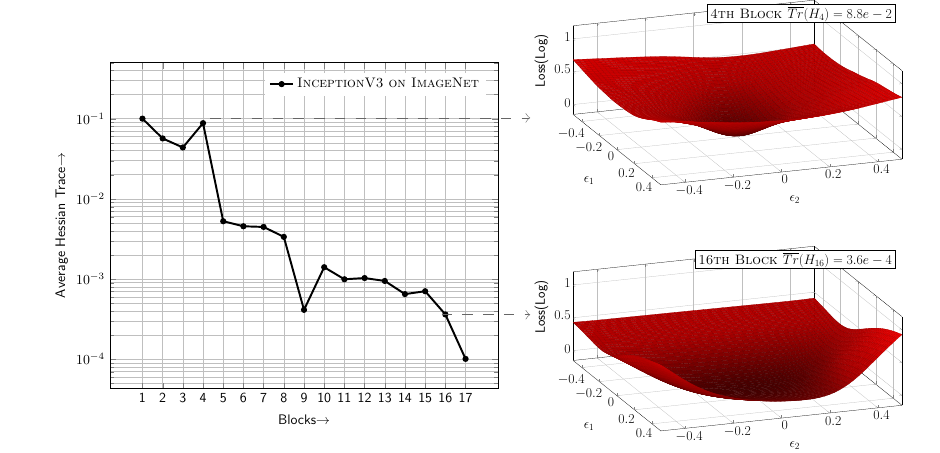}
\includegraphics[width=0.45\textwidth,trim=0.2in 0in 0.in 0.0in, clip]{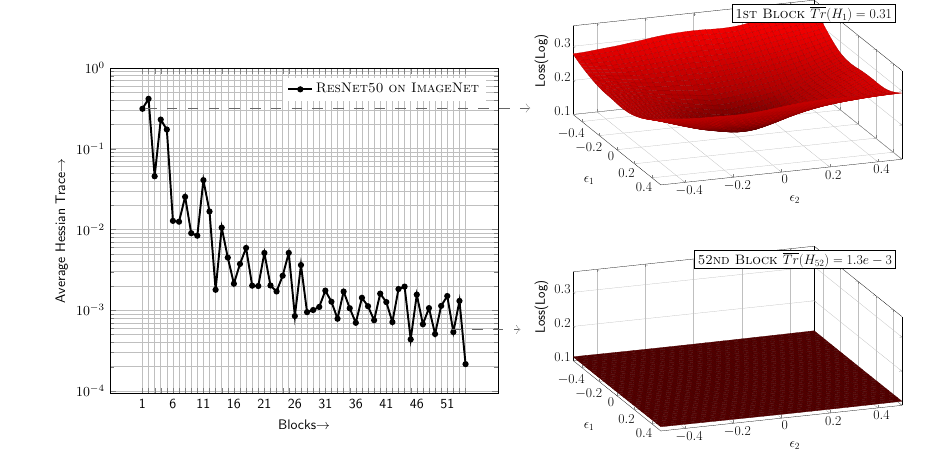}
\caption{
\footnotesize{
Average Hessian trace of different blocks in Inception-V3 and ResNet50 on ImageNet, along with the loss landscape of the block 4 and 16 in Inception-V3 (block 1 and 52 in ResNet50). 
As one can see, the average Hessian trace is significantly different for different blocks. 
We use this information to determine the quantization precision setting, i.e., we assign higher bits for blocks with larger average Hessian trace, and fewer bits for blocks with smaller average Hessian trace. 
}
}
\label{fig:resnet_inception_eigs_surface}
\end{figure}
%%%%%%%%%%%%%%%%%%%%%%%%%%%%%%%%%%%%%%%%%%%%%%%%%%%%%%%%%%%%%%%%%%%%%%%%%%%%%%%%%

\begin{assumption}\label{assumption:1}
Assume that:
\begin{itemize}[noitemsep,topsep=0pt,parsep=0pt,partopsep=0pt,leftmargin=*]
\item 
The model is twice differentiable and has converged to a local minima such that the first and second order optimality conditions are satisfied, i.e., the gradient is zero and the Hessian is positive semi-definite.
\item 
If we denote the Hessian of the $i^{th}$ layer as $H_i$, and its corresponding orthonormal eigenvectors as $v^i_1,~v^i_2,...,v^i_{n_i}$, then the quantization-aware fine-tuning perturbation, $\Delta W_i^* = {\arg\min}_{W_i^* + \Delta W_i^* \in Q(\cdot)} L(W_i^* + \Delta W_i^*)$, satisfies
\small
\begin{align}
    \Delta W_i^* &=  \alpha_{bit}v^i_1 + \alpha_{bit}v^i_2 + ... + \alpha_{bit}v^i_{n_i}.
\end{align}
Here, $n_i$ is the dimension of $W_i$, $W_i^*$ is the converging point of $i^{th}$ layer, and $Q(\cdot)$ is the quantization function which maps floating point values to reduced precision values. 
Note that $\alpha_{bit}$ is a constant number based on the precision setting and quantization range.
\normalsize 
\end{itemize}
\end{assumption}
\noindent
Given this assumption, we establish the following lemma.
\\

\begin{lemma}\label{lemma_1}
Suppose we quantize two layers (denoted by $B_1$ and $B_2$) with same amount of perturbation, namely $\norm{\Delta W_1^*}_2^2$ = $\norm{\Delta W_2^*}_2^2$. 
Then, under Assumption~\ref{assumption:1}, we will have:
\small 
\begin{equation}
    \mathcal{L}(W_1^* + \Delta W_1^*, W_2^*, \cdots, W_L^*) \leq \mathcal{L}(W_1^*, W_2^* + \Delta W_2^*, W_3^*, \cdots, W_L^*), \footnote{We will leave $\mathcal{L}(W_i^*; W_1^*,\cdots W_{i-1}^*, W_{i+1}^* \cdots, W_L^*)$ as $\mathcal{L}(W_i^*)$ without confusion. }
\end{equation}
\normalsize
if
\small 
\begin{equation}
    \frac{1}{n_1}Tr(\nabla^2_{W_1}\mathcal{L}(W_1^*)) \leq \frac{1}{n_2}Tr(\nabla^2_{W_2}\mathcal{L}(W_2^*)).
\end{equation}
\normalsize
\end{lemma}

\begin{proof}
Denote the gradient and Hessian of the first layer as $g_1$ and $H_1$, correspondingly. By Taylor's expansion, we have:
\small 
\begin{equation*}
    \mathcal{L}(W^*_1 + \Delta W^*_1) = \mathcal{L}(W^*_1) + g_1^T\Delta W_1^* + \frac{1}{2}\Delta {W^*_1}^T H_1 \Delta W^*_1 = \mathcal{L}(W^*_1) + \frac{1}{2}\Delta {W^*_1}^T H_1 \Delta W^*_1.
\end{equation*}
\normalsize
Here, we have used the fact that the gradient at the optimum point is zero and that the loss function is locally convex. 
Also note that $\mathcal{L}(W_1^*)=\mathcal{L}(W_2^*)$ since the model has the same loss before we quantize any layer.
Based on the assumption, $\Delta W^*_1$ can be decomposed
by the eigenvectors of the Hessian.
As a result we have:
\begin{equation*}
    \Delta {W^*_1}^TH_1\Delta {W^*_1} = \sum_{i=1}^{n_1} {\alpha_{bit,1}^2} {v^1_i}^TH_1v^1_i = \alpha_{bit,1}^2\sum_{i=1}^{n_1}\lambda^1_{i},
\end{equation*}
where $(\lambda^1_{i},v^1_i)$ is the corresponding eigenvalue and eigenvector of Hessian.
Similarly, for the second layer we will have:
$\Delta {W^*_2}^TH_2\Delta {W^*_2} = {\alpha_{bit,2}^2}\sum_{i=1}^{n_2}\lambda^2_{i}$, where $\lambda^2_{i}$ is the $i^{th}$ eigenvalue of $H_2$.
Since $\norm{\Delta W_1^*}_2$ = $\norm{\Delta W_2^*}_2$, we have $\sqrt{n_1}\alpha_{bit,1}$ = $\sqrt{n_2}\alpha_{bit,2}$. Therefore, we have:
\begin{equation*}
    \mathcal{L}(W^*_2 + \Delta W^*_2) - \mathcal{L}(W^*_1 + \Delta W^*_1) = \alpha_{bit,2}^2n_2(\frac{1}{n_2}\sum_{i=1}^{n_2}\lambda^2_{i} - \frac{1}{n_1}\sum_{i=1}^{n_1}\lambda^1_{i}) \geq 0.
\end{equation*}
It is easy to see that the lemma holds since the sum of eigenvalues equals to the trace of the matrix.
\end{proof}

~\\
\indent 
At first, computing the Hessian trace may seem a prohibitive task, as we do not have direct access to the elements of the Hessian matrix. 
Furthermore, forming the Hessian matrix explicitly is not computationally feasible. 
However, it is possible to leverage the extensive literature in Randomized Numerical Linear Algebra (RandNLA) \cite{Mah-mat-rev_BOOK,PCMI_math_of_data_BOOK} which address this type of problem. 
In particular, the seminar works of~\cite{avron2011randomized,bai1996some} have proposed randomized algorithms for fast trace estimation, using so-called matrix-free methods which do not require the explicit formation of the Hessian operator. 
Here, we are interested in the trace of a symmetric matrix $H\in R^{d\times d}$. 
Then, given a random vector $z\in R^{d}$ whose component is i.i.d. sampled Gaussian distribution ($N(0,1)$) (or Rademacher distribution), we have:
\small 
\begin{equation}
    Tr(H) = Tr(HI) = Tr(H\E[zz^T]) = \E[Tr(Hzz^T)] = \E[z^THz],
\end{equation}
\normalsize
where $I$ is the identity matrix. Based on this, the Hutchinson algorithm~\cite{avron2011randomized} can be used to estimate
the Hessian trace:
\small 
\begin{equation}
    Tr(H) \approx \frac{1}{m}\sum_{i=1}^m z_i^THz_i = Tr_{Est}(H).
\end{equation}

\normalsize
We show empirically in ~\S\ref{sec:Hutchinson} that this algorithm has good convergence properties, resulting in trace computation being orders of magnitude faster than training the network itself.

We have incorporated the above approach and computed the average Hessian trace for different
layers of Inception-V3 and ResNet50, as shown in~\fref{fig:resnet_inception_eigs_surface}. 
As one can see, there is a significant difference between average Hessian trace for different layers. 
To better illustrate this, we have also plotted the loss landscape of Inception-V3 and ResNet50 by perturbing the pre-trained model along the first and second eigenvectors of the Hessian for each layer. 
It is clear that different layers have significantly different ``sharpness.'' 
For instance, the fourth block of Inception-V3 is very sensitive, and thus it needs to be kept at higher bit precision, whereas the 16th block exhibits a very ``flat'' loss landscape and can be quantized more aggressively. 
(In Appendix~\ref{sec:extra_results}, we also show the average Hessian trace for different blocks of SqueezeNext and RetinaNet, as well as their corresponding loss landscape; see \fref{fig:squeezenext_retinanet_eigs_surface}.)

%%%%%%%%%%%%%%%%%%%%%%%%%%%%%%%%%%%%%%%%%%%%%%%%%%%%%%%%%%%%%%%%%%%%%%%%%%%%%%%%%
\begin{figure}[t]
\centering
\includegraphics[width=0.8\textwidth,trim=0.2in 0in 0.in 0.in, clip]{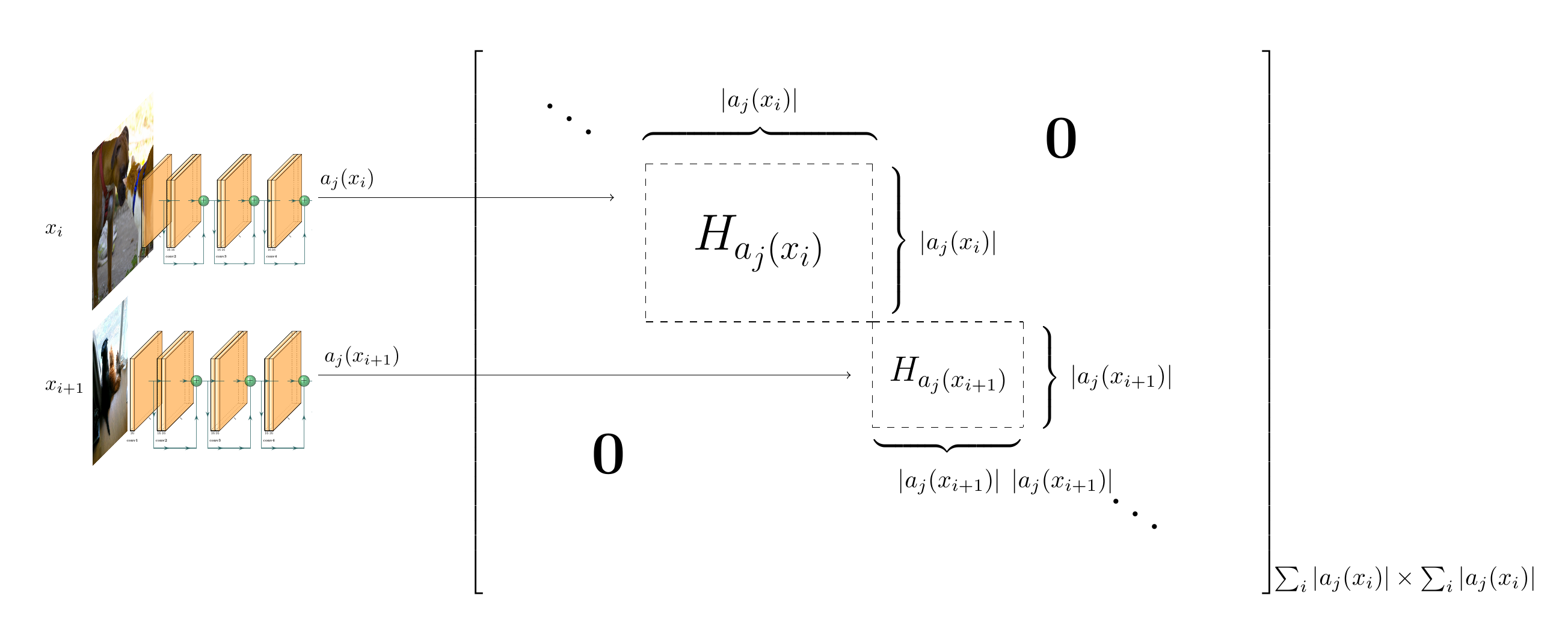}
\caption{
\footnotesize{
Illustration of the structure of Hessian w.r.t to activations ($H_{a_j}$). It is evident that different sized inputs $x_i$ will produce different sized blocks $H_{a_j(x_i)}$ which appear on the diagonal of $H_{a_j}$. 
}
}
\label{fig:hessian_act}
\end{figure}
%%%%%%%%%%%%%%%%%%%%%%%%%%%%%%%%%%%%%%%%%%%%%%%%%%%%%%%%%%%%%%%%%%%%%%%%%%%%%%%%%

\subsection{Mixed Precision Activation}
\label{activation_trace}
The above analysis is not restricted to weights, and in fact it can be extended to mixed-precision activation quantization. 
In \S~\ref{sec:results}, we will show that this is particularly useful for tasks such as object detection.
The theoretical results remain the same.
The only difference is that here the Hessian is with respect to activations instead of model parameters (i.e., second order derivative of the loss w.r.t. activations).
However, computing the Hessian trace w.r.t. each layer's activations is not straight-forward, and even a na\"{\i}ve matrix-free approach can have a very high computational cost.
In the matrix-free Hutchinson algorithm, we need the result of the following Hessian-vector product to compute the Hessian trace:
\begin{equation}
    z^TH_{a_j}z = z^T \left( \nabla^2_{a_j} \frac{1}{N}\sum_{i=1}^N f(x_i, y_i, \theta) \right) z ,
\end{equation}
where $a_j$ is the activations of the $j^{th}$ layer. 
Here, $H_{a_j} \in \R^{(\sum_{i=1}^N | a_j(x_i)|) \times (\sum_{i=1}^N | a_j(x_i)|)}$, where $| a_j(x_i)|$ is the size of the activation of the $j^{th}$ layer for $i^{th}$ input. 
This is because $a_j$ is a concatenation of $a_j(x_i), \forall i$.
See \fref{fig:hessian_act} for the illustration of the matrix $H_{a_j}$ and its shape.
Note that not only is it prohibitive to compute this Hessian matrix, the Hessian-vector product is also infeasible since even generating the random vectors $z \in \R^{\sum_{i=1}^N | a_j(x_i)|}$ is prohibitive, let alone computing its product with $H_{a_j}$.
Furthermore, note that $a_j$ depends on $x_i$, and that for many tasks, such as object detection on Microsoft COCO dataset, $x_i$ does not have a fixed size.
As a result, the activation size of each layer depends on the input data and is not fixed, which further complicates computing Hessian trace w.r.t. activations.

However, the Hessian w.r.t. activations, $H_{a_j}$, has a very interesting structure. 
It is in fact a block-diagonal operator w.r.t. each input data.
That is, $H_{a_j(x_i)}=\nabla^2_{a_j(x_i)}\frac{1}{N}f(x_i, y_i, \theta)$, $H_{a_j}$ is block diagonal, with $H_{a_j(x_i)}$ being the blocks, as illustrated in \fref{fig:hessian_act}.
This observation is simply due to the fact that different inputs are independent of each other.
To show this more formally, let $x_n, x_m$ be two different inputs to the network, and let $g_{a_j(x_m)} = \nabla_{a_j} f(x_m, y_m, \theta)$. 
Notice that
\begin{equation}
    \nabla_{a_j(x_n)} g_{a_j(x_m)} = \mathbf{0},
\end{equation}
since $g_{a_j(x_m)}$ only depends on $x_m$, but not $x_n$.
This observation allows us to compute the Hessian-trace for the layer's activations \textit{for one input} at a time, and then average the resulting Hessian-traces of each block diagonal part, i.e., 
\begin{equation}
    z^T H_{a_j} z = \frac{1}{N} \sum_{i=1}^N z_i^T H_{a_j(x_i)} z_i,
\end{equation}
where $z_i$ is the corresponding components of $z$ w.r.t. the $i^{th}$ input, i.e. $x_i$.
We note that usually this trace computation converges very fast, and it is not necessary to average over the entire dataset.
See~\fref{fig:retinanet_trace} in Appendix for more details.

%%%%%%%%%%%%%%%%%%%%%%%%%%%%%%%%%%%%%%%%%%%%%%%%%%%%%%%%%%%%%%%%%%%%%%%%%%%%%%%%%%%%
\begin{figure}[t]
\centering
\includegraphics[width=.90\textwidth]{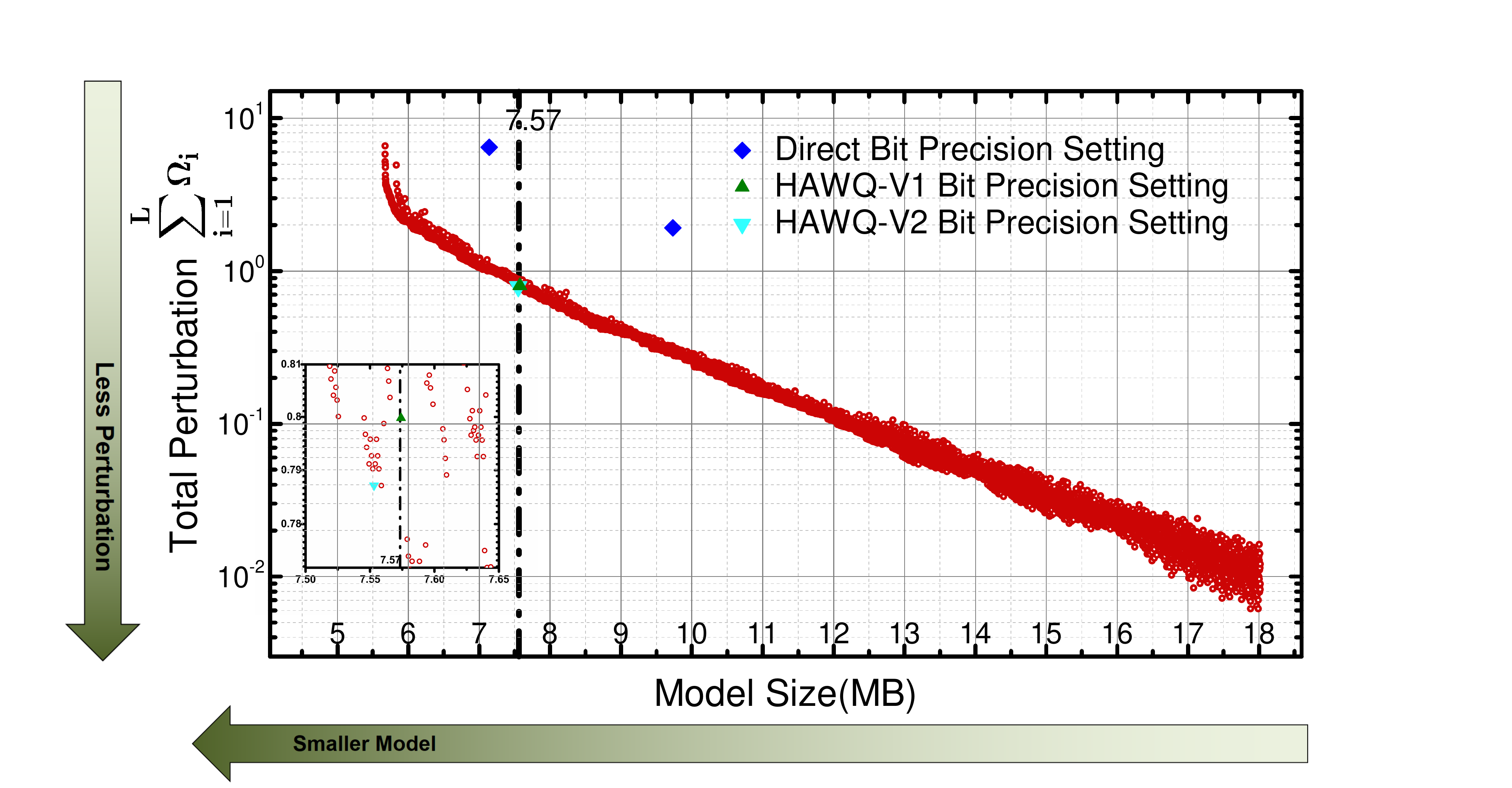}
\caption{
  Pareto Frontier: The trade-off between model size and the sum of $\Omega$ metric (of Eqn.~(\ref{eqn:define_O})) in Inception-V3. 
  Here, $L$ is the number of blocks in the model, and each point in the figure stands for a specific bit precision setting. We show the bit precision setting used in Direct quantization as well as \PREV. To achieve fair comparison, we set constraint on \OURS to have the same model size as \PREV.
  }
  \label{fig:omega_size}
\end{figure}
%%%%%%%%%%%%%%%%%%%%%%%%%%%%%%%%%%%%%%%%%%%%%%%%%%%%%%%%%%%%%%%%%%%%%%%%%%%%%%%%%%%%

\subsection{Weight Bit Selection}
An important limitation of relative sensitivity analysis is that it does not provide
the specific bit precision setting for different layers. This is true
even if we use the average Hessian trace, instead of the top Hessian eigenvalue.
For example, we show the average Hessian trace for different blocks of
Inception-V3 in~\fref{fig:resnet_inception_eigs_surface}. We can clearly see that 
block 1 to block 4 have the largest average Hessian trace, and block 9 or block 16 have orders of magnitude smaller average Hessian trace.
Therefore, while we know that the first four blocks are more sensitive than block 9 or block 16, and thus would benefit from higher number of bits, we still cannot get a specific bit precision setting.

Denote by $\mathcal{B}$ the set of all admissible bit precision settings that satisfy the relative sensitivity analysis based on the average Hessian trace discussed above.
Compared to the original exponential search space, applying the sensitivity constraint makes the cardinality (size) of $\mathcal{B}$ significantly smaller. 
As an example, the original mixed-precision search space for ResNet50 is $4^{50} \approx 1.3\times 10^{30}$ if bit-precisions are chosen among $\{1,2,4,8\}$. 
Using the Hessian-trace sensitivity constraint significantly reduces this search space\footnote{Details on how to calculate the size of $|\mathcal{B}|$ are included in Appendix~\ref{sec:search_space}} to $|\mathcal{B}|=2.3\times 10^{4}$. 
However, this search space is still prohibitively large, especially for deeper models such as ResNet152. 
In the \PREV paper~\cite{dong2019hawq}, the authors manually chose the bit precision among this reduced search space, but this manual selection is undesirable.

We found that this problem can be efficiently addressed using a Pareto frontier approach. 
The main idea is to sort each candidate bit-precision setting in $\mathcal{B}$ based on the total second-order perturbation that they cause, according to the following metric:
\begin{equation}\label{eqn:define_O}
    \Omega = \sum_{i=1}^{L} {\Omega_i} = \sum_{i=1}^{L} \overline{Tr}(H_i) \cdot \|Q(W_i)-W_i\|_2^2,
\end{equation}
where $i$ refers to the $i^{th}$ layer, L is the number of layers in the model,
$\overline{Tr}(H_i)$ is the average Hessian trace, and $\|Q(W_i)-W_i\|_2$ is the $L_2$ norm of quantization perturbation. 
The intuition is that a bit precision setting with minimal second-order perturbation to
the model should lead to good generalization after quantization-aware fine-tuning. 
Given a target model size, we sort the elements of $\mathcal{B}$ based on
their $\Omega$ value, and we choose the bit precision setting with minimal $\Omega$.
While this approach cannot theoretically guarantee the best possible
performance, we have found that in practice it can generate bit precision settings that exceed all state-of-the-art results, and it removes the manual precision selection process used in \PREV~\cite{dong2019hawq}.\footnote{It should be noted that we can compute $\Omega_i$ in negligible time on a single CPU since it does not require quantization-aware fine-tuning. Typically, it takes less than 1 second to compute $\Omega$ value for $10^3$ entries in $\mathcal{B}$.}

We show the process for choosing the exact bit precision setting of Inception-V3 in~\fref{fig:omega_size}. 
Each red dot denotes a specific bit precision setting for different blocks of Inception-V3 that satisfy the Hessian trace constraint. 
For each target model size, \OURS chooses the bit precision setting with minimal $\Omega$ value. 
With green triangles, we have also denoted the bit precision setting that was manually selected in the \PREV paper~\cite{dong2019hawq}. 
The automatic bit precision setting of \OURS exceeds the accuracy of \PREV, as will be discussed in the next section.

\section{Results}
\label{sec:results}

%~\\
%\indent 

In this section, we first analyze the convergence of the Hutchinson algorithm and the speed of Hessian trace calculation with the Hutchinson algorithm. 
Then, we show state-of-the-art quantization results achieved by \OURS on both ImageNet for classification and Microsoft COCO for object detection. 
We emphasize that all of these results are achieved without any AutoML based search or manaul bit-precision selection.

\subsection{Hutchinson}\label{sec:Hutchinson}
\begin{figure}[h]
\centering
\includegraphics[width=0.45\textwidth,trim=0in 0in 0.in 0.in, clip]{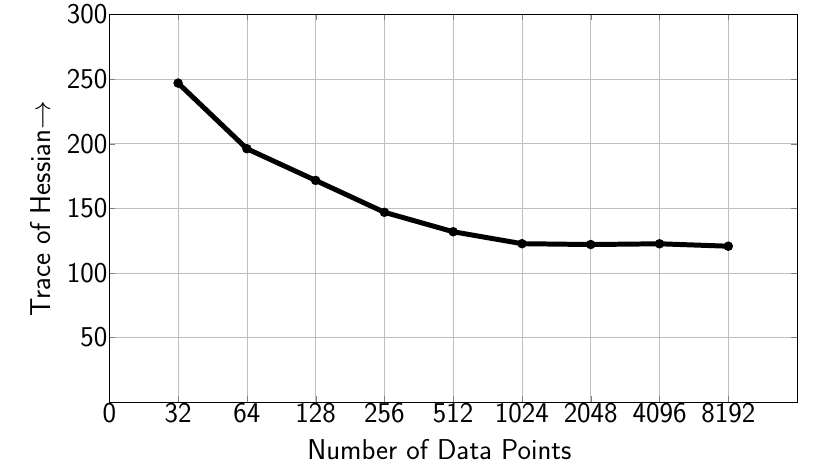}
\includegraphics[width=0.45\textwidth,trim=0in 0in 0.in 0.0in, clip]{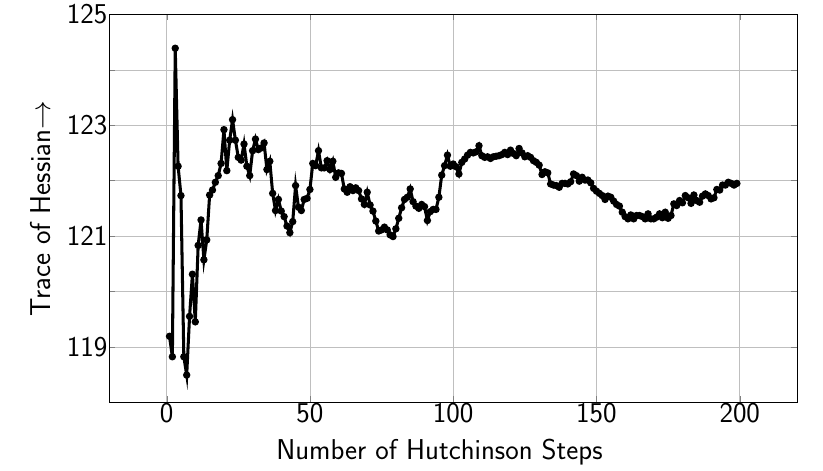}
\caption{
\footnotesize{
Relationship between the convergence of Hutchinson and the number of data points (Left) as well as the number of steps (Right) used for trace estimation on block 21 in ResNet50. 
}
}
\label{fig:resnet_trace}
\end{figure}

In~\fref{fig:resnet_trace}, we show how the convergence of the Hutchinson algorithm is related to the number of data points and the number of Hutchinson steps used for trace estimation. 
It can be clearly seen that the trace estimation converges rapidly as we increase the number of data points over 512, over which the sub-sampled Hessian is computed (i.e. $N_B$).
Moreover, we can see that 50 Hutchinson steps are sufficient to achieve an accurate approximation with low variance. 
Based on the convergence analysis, we are able to calculate all the average Hessian traces, shown in~\fref{fig:resnet_inception_eigs_surface}, corresponding to 54 blocks in a ResNet50 model, within 30 minutes (33s per block on average) using 4 GPUs. 
The Hutchinson algorithm, in addition to the automatic bit precision and quantization order selection, makes \OURS a significantly faster algorithm than previous reinforcement learning based algorithms~\cite{wang2018haq}.

% -----------------------------------------------------------------0
% -----------------------------------------------------------------

\subsection{ImageNet} 

%~\\
%\indent 

As shown in~\tref{tab:inception_compress_ratio}, we first apply \OURS on Inception-V3~\cite{szegedy2016rethinking}. Direct quantization of InceptionV3 (i.e., without use of second-order information), results in $7.69\%$ accuracy degradation. 
Using the approach proposed in~\cite{jacob2018quantization} results in more than $2\%$ accuracy drop, even though it uses higher bit precision.
However, \PREV~\cite{dong2019hawq} results in a $2\%$ accuracy gap with a compression ratio of $12.04\times$, both of which are better than previous work~\cite{jacob2018quantization,park2018value}.
Although \PREV uses second-order information to obtain a relative order of quantization precision for each block, the exact bit precision needed to be selected manually. 
In contrast, \OURS can automatically generate the exact precision setting for the whole network, while being able to achieve better accuracy than \PREV.

\begin{table}[h]
\caption{
Quantization results of Inception-V3 on ImageNet. 
We abbreviate 
quantization bits used for weights as ``w-bits,'' 
quantization bits used for activations as ``a-bits,''
top-1 testing accuracy as ``Top-1,'' and 
weight compression ratio as ``W-Comp.''
Furthermore, we compare \OURS with direct quantization method without using Hessian (``Direct'')
and Integer-Only~\cite{jacob2018quantization}. Here ``MP'' refers to mixed-precision quantization.
Compared to~\cite{jacob2018quantization,park2018value}, we achieve higher compression ratio with higher
testing accuracy.
}
\small
\setlength\tabcolsep{2.35pt}
\label{tab:inception_compress_ratio}
\centering
\begin{tabular}{lcccccccccccccc} \toprule

Method                                          & w-bits    & a-bits  & Top-1        &  W-Comp          & Size(MB)  
\\ 
\midrule
\hc  Baseline                                   & 32   & 32 & 77.45       & 1.00$\times$    &  91.2        \\ 
\midrule
\ha  Integer-Only~\cite{jacob2018quantization}  & 8    & 8  & 75.40       & 4.00$\times$    &  22.8        \\ 
\ha  Integer-Only~\cite{jacob2018quantization}  & 7    & 7  & 75.00       & 4.57$\times$    &  20.0       \\ 
\ha  RVQuant~\cite{park2018value}  & 3\MP    & 3\MP  & 74.14       & 10.67$\times$    &  8.55        \\
\ha  Direct                                 & 2\MP    & 4\MP  & 69.76       & 15.88$\times$   &  5.74     
\\
\ha \PREV~\cite{dong2019hawq}               & 2\MP    & 4\MP  & 75.52  & 12.04$\times$   &  \bf{7.57}        \\ 
\midrule
\hc \OURS                                   & 2\MP    & 4\MP  & \bf{75.68}  & 12.04$\times$   &  \bf{7.57}        \\ 
     \bottomrule 
\end{tabular}
\end{table}

We also show \OURS results on ResNet50~\cite{he2016deep}, and compare \OURS with other popular quantization methods~\cite{zhou2016dorefa, choi2018pact, zhang_2018_lqnets, han2015deep, wang2018haq, dong2019hawq} in~\tref{tab:resnet50_compress_ratio}. 
It should be noted that~\cite{zhou2016dorefa, choi2018pact, zhang_2018_lqnets, han2015deep} followed traditional quantization rules which set the precision for the first and last layer to 8-bit, and quantized other layers to an identical precision.
Both~\cite{wang2018haq, dong2019hawq} are mixed-precision quantization methods. 
Also,~\cite{wang2018haq} uses reinforcement learning methods to search for a good precision setting, while \PREV uses second-order information to guide the precision selection as well as the block-wise fine-tuning. 
\PREV achieves the state-of-the-art accuracy $75.48\%$ with a 7.96MB model size. 
Keeping model size the same, \OURS can achieve $75.76\%$ accuracy without any heuristic knowledge and manual efforts.

\begin{table}[h]
\caption{
Quantization results of ResNet50 on ImageNet.
We show results of state-of-the-art methods~\cite{zhou2016dorefa, choi2018pact, zhang_2018_lqnets, han2015deep}.
We  also  compare  with  the  recent AutoML approach of~\cite{wang2018haq}.
Compared to~\cite{wang2018haq}, we achieve higher compression ratio with higher
testing accuracy. Also note that \cite{zhou2016dorefa, choi2018pact, zhang_2018_lqnets} use 8-bit for first and last layers.
}
\small
\setlength\tabcolsep{2.35pt}
\label{tab:resnet50_compress_ratio}
\centering
\begin{tabular}{lcccccccccccccc} \toprule
Method                                          & w-bits    & a-bits  & Top-1        &  W-Comp          & Size(MB)  
\\ 
\midrule
\hc  Baseline                                   & 32   & 32 & 77.39       & 1.00$\times$    &  97.8        \\ 
\midrule
\ha  Dorefa~\cite{zhou2016dorefa}  & 2    & 2  & 67.10       & 16.00$\times$    &  6.11        \\ 
\ha  Dorefa~\cite{zhou2016dorefa}  & 3    & 3  & 69.90       & 10.67$\times$    &  9.17       \\ 
\ha  PACT~\cite{choi2018pact}  & 2    & 2  & 72.20       & 16.00$\times$    &  6.11        \\ 
\ha  PACT~\cite{choi2018pact}  & 3    & 3  & 75.30       & 10.67$\times$    &  9.17       \\ 
\ha  LQ-Nets~\cite{zhang_2018_lqnets}     & 3      & 3  & 74.20      & 10.67$\times$  & 9.17 \\ 
\ha Deep Comp.~\cite{han2015deep}               & 3    & MP  & {75.10}  & 10.41$\times$  &  9.36        \\  
\ha HAQ~\cite{wang2018haq}                 & MP   & MP  & {75.30}  & 10.57$\times$  &  9.22         \\ 
\ha \PREV~\cite{dong2019hawq}                                   & 2\MP    & 4\MP  & 75.48  & 12.28$\times$   &  7.96        \\ 
\midrule
\hc \OURS                                   & 2\MP    & 4\MP  & \bf{75.76}  & 12.24$\times$   &  \bf{7.99}        \\
     \bottomrule 
\end{tabular}
\end{table}

We also apply \OURS to quantize deep and highly compact models such as SqueezeNext~\cite{gholami2018squeezenext}. 
We choose the wider SqueezeNext model which has a baseline accuracy of $69.38\%$ with 2.5 million parameters (10.1MB in single precision).
We can see from~\tref{tab:squeezenext_compress_ratio} that direct quantization of SqueezeNext (i.e., without use of second-order information), results in $3.98\%$ accuracy degradation. 
\PREV results in a 1MB model size, with 1.36\% top-1 accuracy drop. 
By applying \OURS on SqueezeNext, we can achieve a $68.38\%$ accuracy with an unprecendented model size of 1.07MB (which is even slightly smaller than \PREV).

\begin{table}[h]
\caption{
Quantization results of SqueezeNext on ImageNet.
We first show results of direct quantization method without using Hessian (``Direct'').
Then we compare \OURS with \PREV, which can compress SqueezeNext to a model 
with an unprecedented 1MB model size with only 1.36\% top-1 accuracy drop. 
By applying \OURS on SqueezeNext, we can achieve even better accuracy $68.38\%$ with even smaller model size than \PREV.
}
\small
\setlength\tabcolsep{2.35pt}
\label{tab:squeezenext_compress_ratio}
\centering
\begin{tabular}{lcccccccccccccc} \toprule
Method              & w-bits    & a-bits  & Top-1        &  W-Comp          & Size(MB)  \\ 
\midrule
\hc  Baseline       & 32    & 32    & 69.38        & 1.00$\times$          &  10.1        \\ 
\midrule
\ha Direct      & 3\MP     & 8     & 65.39   & 9.04$\times$          &  1.12        \\
\ha \PREV~\cite{dong2019hawq}       & 3\MP     & 8     & 68.02   & 9.26$\times$          &  \bf{1.09}        \\
\hc \OURS       & 3\MP     & 8     & \bf{68.38}   & 9.40$\times$          &  \bf{1.07}        \\
     \bottomrule 
\end{tabular}
\end{table}

\subsection{Microsoft COCO}
In order to show the generalization capability of \OURS, we also test object detection task Microsoft COCO 2017~\cite{lin2014coco}.
This contains 118k training images(40k labeled) with 80 object categories. 
RetinaNet~\cite{lin2017focal} is a single stage detector that can achieve state-of-the-art mAP\footnote{Here we use the standard mAP 0.5:0.05:0.95 metric in COCO dataset.} with a very simple network architecture, and it only contains hardware-friendly operations such as convolutions and additions. As shown in~\tref{tab:retinanet}, we use the pretrained RetinaNet with ResNet50 backbone as our baseline model, which can achieve 35.6 mAP with 145MB model size. We first show the result of direct quantization where no Hessian information is used. Even with quantization-aware fine-tuning and channel-wise quantization of weights, directly quantizing weights and activations
in RetinaNet to 4-bit causes a significant 4.1 mAP degradation. FQN~\cite{li2019fully} is a recently proposed quantization method 
which reduces this accuracy gap to 3.1 mAP with the same compression ratio as Direct method.
Using \OURS on mixed-precision weight quantization with uniform 4-bit activations can achieve a state-of-the-art performance of 34.1 mAP, which is 1.6 mAP higher than~\cite{li2019fully} with an even smaller model size.

\begin{table}[h]
\caption{
Quantization results of RetinaNet on Microsoft COCO 2017.
We show results of direct quantization, as well as a state-of-the-art quantization method for object detection~\cite{li2019fully}.
With the same model size, \OURS can outperform previous quantization results by a large margin. 
We also show that \OURS with mixed-precision activations can achieve even better mAP, with a slightly lower activation compression ratio.
}
\small
\setlength\tabcolsep{2.35pt}
\label{tab:retinanet}
\centering
\begin{tabular}{lcccccccccccccc} \toprule
Method                                          & w-bits    & a-bits  & mAP        &  W-Comp     & A-Comp     & Size(MB)  
\\
\midrule
\hc  Baseline                                   & 32   & 32 & 35.6       & 1.00$\times$  &1.00$\times$  &  145        \\ 
\midrule
\ha  Direct                    & 4    & 4  & 31.5       & 8.00$\times$  & 8.00$\times$  &  18.13        \\ 
\ha  FQN~\cite{li2019fully}  & 4    & 4  & 32.5       & 8.00$\times$  & 8.00$\times$  &  18.13        \\ 
\ha \OURS                      & 3\MP    & 4  &  \bf{34.1}  & 8.10$\times$ & 8.00$\times$  &  17.90        \\
\midrule
\hc \OURS                      & 3\MP    & 4\MP  & \bf{34.4}  & 8.10$\times$ & 7.62$\times$  &  17.90        \\
\hc \OURS                      & 3\MP    & 6  & \bf{34.8}  & 8.10$\times$ & 5.33$\times$  &  17.90        \\
     \bottomrule 
\end{tabular}
\end{table}

It should also be noted that we found the activation quantization to be very sensitive for object detection models.
For instance, increasing activation quantization bit precision to 6-bit, results in a 34.8 mAP, which is 0.7 mAP higher, as compared to 34.1 mAP achieved with 4-bit activation.

One might argue that using 6-bit for activation results in higher activation memory. 
This can be a problem for extreme cases such as deploying these models on micro-controllers where every bit counts. 
For these situations, we can use mixed-precision activation. 
This can be performed using the Hessian AWare technique discussed in~\S\ref{activation_trace}, with the same automatic bit-precision selection method using Pareto optimal curve.
As can be seen in~\tref{tab:retinanet}, mixed-precision activation quantization can achieve very good trade-off between accuracy and compression. 
With only marginal change to activation compression ratio,
it can achieve 34.4 mAP, which significantly outperforms uniform 4-bit activation quantization, and is even close to a uniform 6-bit activation quantization.

\section{Conclusions and Future Work}
\label{sec:conclusions}
In this work, we presented several improvements over the basic \PREV method~\cite{dong2019hawq}. 
We performed a theoretical analysis showing that a better sensitivity metric is to use the Hessian trace, instead of just the top Hessian eigenvalue. 
We extended the framework to mixed-precision activation, and we proposed a very efficient method for computing the Hessian trace with respect to activations by using matrix-free algorithms. 
Furthermore, we presented an automatic bit-precision setting to avoid the manual bit selection used in \PREV~\cite{dong2019hawq}. 
We presented state-of-the-art results on image classification for Inception-V3 (75.68\% with 7.57MB model size), ResNet50 (75.76\% with 7.99MB model size) and SqueezeNext (68.38\% with 1MB model size). 
Furthermore, we showed results for object detection task, where we applied \OURS on RetinaNet on Microsoft COCO dataset. 
Our quantized results achieve more than 1.6 mAP higher accuracy than the recently proposed method FQN~\cite{li2019fully}, with an even smaller model size of 17.9MB (as compared to 18.13MB)

Using second-order information has typically been viewed as a mere theoretical tool in machine learning, but our results have shown that this is not the case, and that significant gains can be attained in practice by considering higher order Hessian information.
Similar recent work has also shown promising results when using second-order information for adversarial attacks~\cite{yao2019trust} and the analysis of large batch size training~\cite{yao2018hessian,yao2018large}.
A promising future step for quantization is to use second-order information throughout the training process, encouraging the final converged model (to which quantization is then applied) to have a flatter loss landscape for most or all of the layers. 
This could allow for an even lower bit-precision quantization without accuracy degradation. 
Another important future direction is to extend this analysis for cases where training data is not accessible. 
This is very common in practice due to specific regulations such as privacy constraints.

\section*{Acknowledgments}
This work was supported by a gracious fund from Intel corporation, and in particular
Intel VLAB team. We are also grateful for a gracious fund from Google Cloud, Google TFTC team, as well as  support from the Amazon AWS.
We also thank NVIDIA Corporation with the donation of the Titan Xp GPUs that was partially used for this research.
MWM would also like to acknowledge ARO, DARPA, NSF, ONR, Cray, and Intel for providing partial support of this work.

\bibliographystyle{plain}
{
\bibliography{ref.bib}}

\clearpage
\appendix
\section{Quantization Details}\label{sec:quantization}
During the forward pass, each element in a weight or activation tensor $X$ will be quantized as~follows:
\begin{align*}
X^\prime = \text{Clamp}&(X, q_0, q_{2^{k}-1}), \\
X^I = \lfloor \frac{X^\prime - q_0}{\Delta} \rceil,& \text{ where } \Delta = \frac{q_{2^{k}-1} - q_0}{2^k - 1}, \\
Q(X) = \Delta &X^I + q_0 ,
\end{align*}
where $\lfloor \cdot \rceil$ is the round operator, $\Delta$ is the distance between adjacent quantized points, $X^I$ is a set of integer indices, $[q_0, q_{2^{k}- 1}]$ stands for the quantization range of the floating point tensor, and the function Clamp sets all elements smaller than $q_0$ equal to $q_0$, and all elements larger than $q_{2^{k}- 1}$ to $q_{2^{k}- 1}$. 
It should be noted that $[q_0, q_{2^{k}- 1}]$ can be a subinterval of $[min, max]$, in order to get rid of outliers and better represent the majority of the given tensor. 
During inference, the expensive floating point arithmetic can be replaced 
by efficient integer arithmetic for the matrix multiplication with $X^I$, 
and then followed by a gathered dequantization operation, 
which will significantly accelerate the computation process. Since we use the quantization-aware fine-tuning scheme, in the backward pass, the Straight-Through Estimator (STE)~\cite{bengio2013estimating} is used for computing the gradient for $X$.

\section{Search Space}\label{sec:search_space}
Suppose the number of different mixed-precision settings is $\mathcal{B}$, and the number of different progressive quantization-aware fine-tuning orders is $\mathcal{C}$. 
The whole search space can be written as $\mathcal{B}\times \mathcal{C}$. We have:
\begin{equation*}
    \mathcal{B} = m^{L}.
\end{equation*}
\begin{equation*}
    \mathcal{C} = \sum_{i=1}^{L} {i!\times S(L,i)} \rightarrow L!.
\end{equation*}
where $m$ is the number of quantization precision options, $L$ is the number of layers in a given model, $S(L,i)$ stands for Stiring numbers of the second kind, which have a growth speed between $O(L!)$ and $O(L^L)$. 
In the case of layer-wise fine-tuning, where only one layer can be fine-tuned at a time, $\mathcal{C}$ degrades to $L!$.

Given two layers $B_i$ and $B_j$ with average Hessian trace $Tr(B_i)/n_i > Tr(B_j)/n_j$, if we set quantization precision $q_i\geq q_j$, then based on that, we are able to order all $L$ layers in the model according to their average Hessian trace. 
Considering the situation that $j$ precision options are used out of total number $m$, the mixed-precision problem can be reduced to an integer partition problem, namely, to partition the ordered layers into $j$ different groups, which results in ${j-1}\choose {L-1}$ possible solutions. 
Since there are $j\choose m$ different combinations of the $j$ precision options, the total size of search space is $\sum_{j=1}^{m} {({j\choose m} \cdot {{j-1}\choose {L-1}})}$.

\section{Extra Results}\label{sec:extra_results}

% -----------------------------------------------------------------
\begin{figure}[h]
\centering
\includegraphics[width=0.45\textwidth,trim=0.2in 0in 0.in 0.0in, clip]{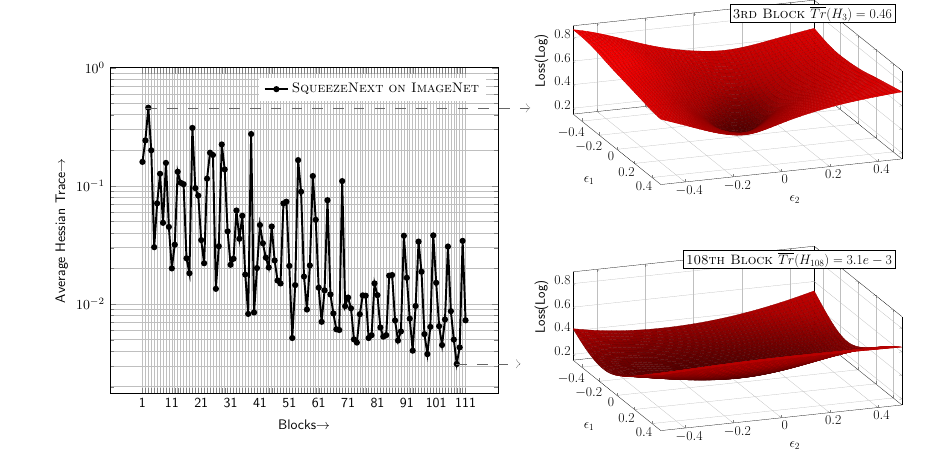}
\includegraphics[width=0.45\textwidth,trim=0.2in 0in 0.in 0.in, clip]{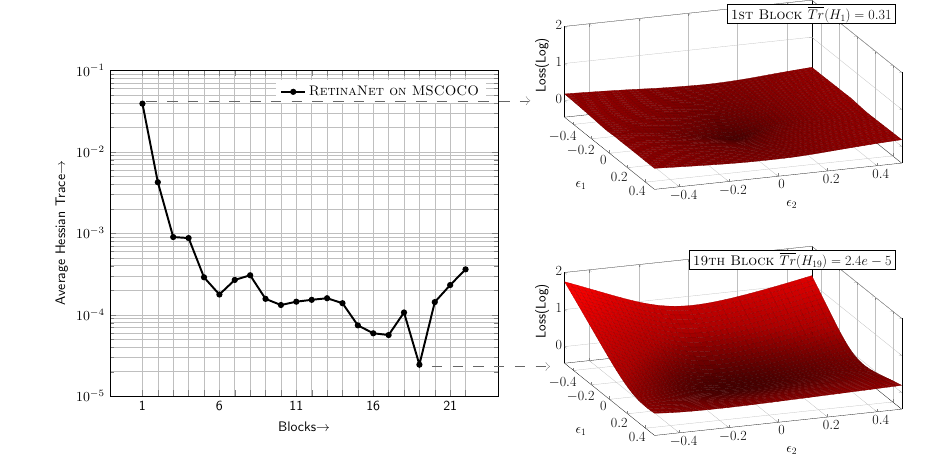}
\caption{
\footnotesize{
Average Hessian trace of different blocks in SqueezeNext and RetinaNet, along with the loss landscape of block 3 and 108 in SqueezeNext, and block 1 and 19 in RetinaNet. It should be noted that block 1 to block 17 in RetinaNet are the ResNet50 backbone, block 18 to block 20 are FPN, and block 21 and block 22 are classification and regression head, respectively. 
As one can see, the average Hessian trace is significantly different for different blocks. We assign higher bits for blocks with larger average Hessian trace, and fewer bits for blocks with smaller average Hessian trace. 
For reference, in~\fref{fig:resnet_inception_eigs_surface} we showed a similar plot but for Inception-V3 and ResNet-50.
}
}
\label{fig:squeezenext_retinanet_eigs_surface}
\end{figure}
% -----------------------------------------------------------------

% -----------------------------------------------------------------
\begin{figure}[h]
\centering
\includegraphics[width=0.36\textwidth,trim=0in 0.02in 0in 0.0in, clip]{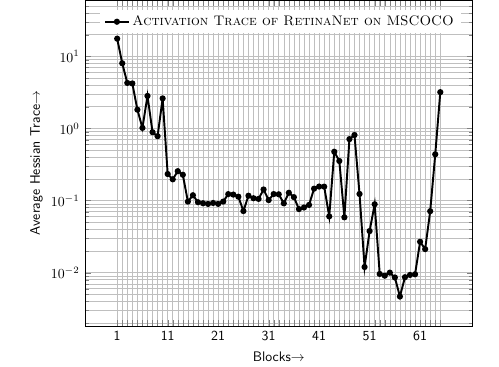}
\includegraphics[width=0.45\textwidth,trim=0in 0in 0.in 0.in, clip]{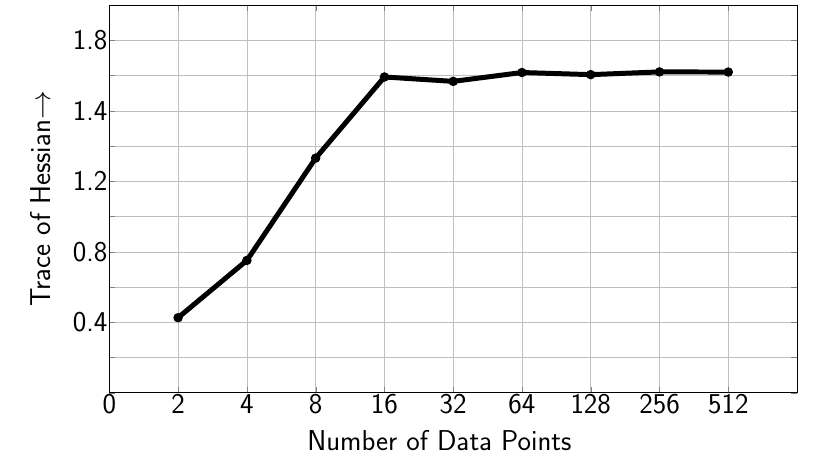}
\caption{
\footnotesize{
(Left) Average Hessian trace w.r.t. activations in RetinaNet. As we can see, the average Hessian trace varies significantly across activations of different blocks. 
We use this information to perform mixed-precision activation quantization as discussed in\S~\ref{activation_trace}.
(Right) we show the relationship between the convergence of Hutchinson and the number of data points used for trace estimation on block 5 in RetinaNet.
We used 128 data points with 50 Hutchinson steps to plot the left figure.
}
}
\label{fig:retinanet_trace}
\end{figure}
% -----------------------------------------------------------------

\end{document}